\documentclass[11pt,a4paper]{article}
\usepackage{times,latexsym}
\usepackage{url}
\usepackage[T1]{fontenc}

\usepackage[acceptedWithA]{tacl2021v1}
\usepackage[]{tacl2021v1}

\usepackage{xspace,mfirstuc,tabulary}

\newif\iftaclinstructions
\taclinstructionsfalse 
\iftaclinstructions
\renewcommand{\confidential}{}
\renewcommand{\anonsubtext}{(No author info supplied here, for consistency with
TACL-submission anonymization requirements)}
\newcommand{\instr}
\fi

\iftaclpubformat 

\else

\fi

\usepackage{enumitem}
\usepackage{amsmath}
\usepackage{multirow}
\usepackage{graphicx}
\usepackage{ulem}
\usepackage{booktabs}
\usepackage{amsfonts} 
\renewcommand{\emph}{\textit}
\title{Bird-Eye Transformers for Text Generation Models}

\author{%
  Lei Sha, Yuhang Song, Yordan Yordanov, Tommaso Salvatori, Thomas Lukasiewicz\\
  Department of Computer Science, University of Oxford, Oxford, UK \\
  \texttt{firstname.lastname@cs.ox.ac.uk}\\
}

\begin{document}

\maketitle 

\begin{abstract}

Transformers have become an indispensable module for text generation models since their great success in machine translation. 
Previous works attribute the~success of transformers to the query-key-value dot-product attention,
which provides a robust inductive bias by the fully connected token graphs.
However, we found that self-attention has a severe limitation.
When predicting the $(i\,{+}\,1)$-th token, self-attention only takes the $i$-th token as an information collector, and it tends to give a high attention weight to those tokens similar to itself. Therefore, most of the historical information that occurred before the $i$-th token is not taken into consideration.
Based on this observation, in this paper, we propose a new architecture, called \textbf{bird-eye transformer}~(\textbf{BET}), which goes one step further to improve the performance of transformers by reweighting self-attention to encourage it to focus more on important historical information.
We have conducted experiments on multiple text generation tasks, including machine translation (2 datasets) and language models (3 datasets).
These experimental~results show that our proposed model achieves a better performance than the baseline transformer
architectures on~all~datasets. The code is released at: \url{https://sites.google.com/view/bet-transformer/home}.
  
\end{abstract}
\section{Introduction}

The successful application of transformers~\cite{vaswani2017attention} in machine translation shows that it is a much better choice for sequence modeling than auto-regressive architectures like RNNs~\cite{rumelhart1986learning} and LSTMs~\cite{hochreiter1997long}. The core of transformers is self-attention, which is a dot-product-based token-by-token correlation computation module. Compared to previous popular auto-regressive architectures, self-attention directly builds the connection between tokens, which also captures long-range dependencies. 

However, we found that self-attention has some severe disadvantages, namely,  the self-attention modules focus too much on the current token and fails to provide specific attention to ``high-level'' historical tokens. The ``high-level'' tokens refer to the type of token that can influence other tokens over a long distance, for example,  the tokens located closely to the to-be-predicted token in the dependency parse tree. For example, in sentence ``\textit{...cat catch a mouse...}'', if the current token is ``\textit{a}'' and we are predicting the token ``\textit{mouse}''. In the self-attention module, ``\textit{a}'' is taken as an information collector to compute the attention weights with other tokens by the dot product. This leads to the fact that the next token ``\textit{mouse}'' is mostly predicted by the information of the token  ``\textit{a}'', while some more important information that occurred before are not being considered enough  (such as ``\textit{cat}'' and  ``\textit{catch}'').

To tackle the above disadvantages of self-attention, in this paper, we
propose to encourage the attention weights to focus more on the ``high-level'' tokens by some syntax guidance. We propose a novel architecture called \emph{bird-eye transformer} (\emph{BET}) to provide transformers with a bird-eye view of all the historical tokens.  \emph{BET} has two alternative architectures to achieve this goal: (1) A syntax-guided transformer architecture, called \emph{BET(SG)}: This architecture takes some syntax hints from the dependency parse tree and use such hints to reweight the self-attention's dot-product matrix.
(2) A syntax-guidance-free transformer architecture, called \emph{BET(SF)}: We don't use any syntax hints in this architecture. To provide a bird-eye view, we first reweight the self-attention's dot-product matrix towards attending to high-level tokens. We achieve this using a function that  decides which tokens are high-level tokens according to the input sequence and the self-attention's output sequence. The architecture is expected to induce the high-level information by itself.

In addition, we show that the attention weights on the current token (in the dot-product matrix) should be split to other historical tokens. This modification will not lead to a loss of the current token’s information, because it will be added back in the residual connection~\cite{he2016deep} part of the transformer. 

 Finally, the refined self-attention matrix is obtained by applying the \texttt{Softmax} on the reweighted dot-product matrix.
The main contributions of this paper are briefly summarized as follows:
\begin{itemize}[leftmargin=*]
    \item We point out the severe disadvantages in the self-attention of transformers, and we report on detailed experimental results that prove these disadvantages.
    \item We propose a novel bird-eye transformer architecture, which refines the self-attention in transformers to provide it with a bird-eye view of the historical information in sequence modeling.
    \item We conduct experiments on multiple tasks and compare against several natural baselines to prove the effectiveness of our proposed bird-eye transformer.
\end{itemize}

\section{Background}

Transformers were proposed by \newcite{vaswani2017attention} for machine translation, using a series of transformer blocks in the encoder and the decoder part.  Each transformer block contains a self-attention layer, a feed-forward layer,  multiple skip-connections, and layer normalizations.  Self-attention is the core component of transformers. Given an input sequence $X=(x_1,\ldots,x_n)$ of token representations, we first use three linear transformations to map $X$ into three matrices: queries $Q$, keys $K$, and values $V$. Then, self-attention is calculated as follows:
\begin{equation}\label{eq:selfattention}
\begin{small}
\begin{aligned}
    &Q=XW_Q,\quad K=XW_K,\quad V=XW_V,\\
    &D={QK^\top}/{\sqrt{d}},\quad A(Q,K,V)=\texttt{Softmax}(D)V,\\
\end{aligned}
\end{small}
\end{equation}
where $d$ is the vector dimension of each token's representation, $W_Q$, $W_K$, and $W_V$ are all trainable parameters, and $D$ is the dot-product matrix. 

Each transformer block has two sub-blocks, one is for self-attention,  and the other is a feed-forward layer.
The results of both the self-attention and the feed-forward operation are then added together with inputs as a residual connection~\cite{he2016deep}, followed by layer normalization:
\begin{align}
    X'&=\texttt{LayerNorm}\big(X + A(Q,K,V)\big),\label{eq:tf1}\\
H &=\texttt{LayerNorm}\big(X' + F\!F\!L(X')\big),\label{eq:tf2}
\end{align}
where $F\!F\!L$ stands for feed-forward layer.

In text generation, when the transformer block is used as decoder, we need to multiply the self-attention matrix with a triangular mask to prevent each token from attending to subsequent tokens.

\section{Approach}

\subsection{Motivation}
Intuitively, the self-attention module should focus on some ``high-level'' tokens instead of  focusing too much on the current token.  

Assume that we are predicting the $(i+1)$-th token relative to the previous tokens $x_0,\ldots,x_i$. According to Eq.~\eqref{eq:selfattention},  the $i$-th row of dot-product matrix $D$ is calculated as:
\begin{equation}\label{eq:sim}
    \resizebox{\linewidth}{!}{$D_i = \big[\frac{1}{\sqrt{d}}x_{i}W_QW_K^\top x_{0}^\top, \ldots,  \frac{1}{\sqrt{d}}x_{i}W_QW_K^\top x_{i}^\top, [M], \ldots,  [M] \big]$},
\end{equation}
where ``[M]'' stands for the masked out elements  that correspond to future tokens.
The dot-product $D$ is used for measuring the relevance between tokens. Intuitively, since a token is always more similar to itself than other tokens, the $i$-th item  $x_iW_QW_K^\top x_i$ is expected to be the largest in Eq.~\eqref{eq:sim}. 

However, in fact, the $i$-th token is not necessary to attend to itself, because in the transformer architecture, the information of the $i$-th token can be added to the collected feature vector by the residual connection~\cite{he2016deep}, as shown in Eq.~\eqref{eq:tf1}. If we mask out the diagonal attention values of the self-attention matrix and split the attention values to  the historical tokens, then the historical tokens will obtain more attention and the current token's attention is also kept by the residual connection.

\begin{figure}
    \centering
    \includegraphics[width=\linewidth]{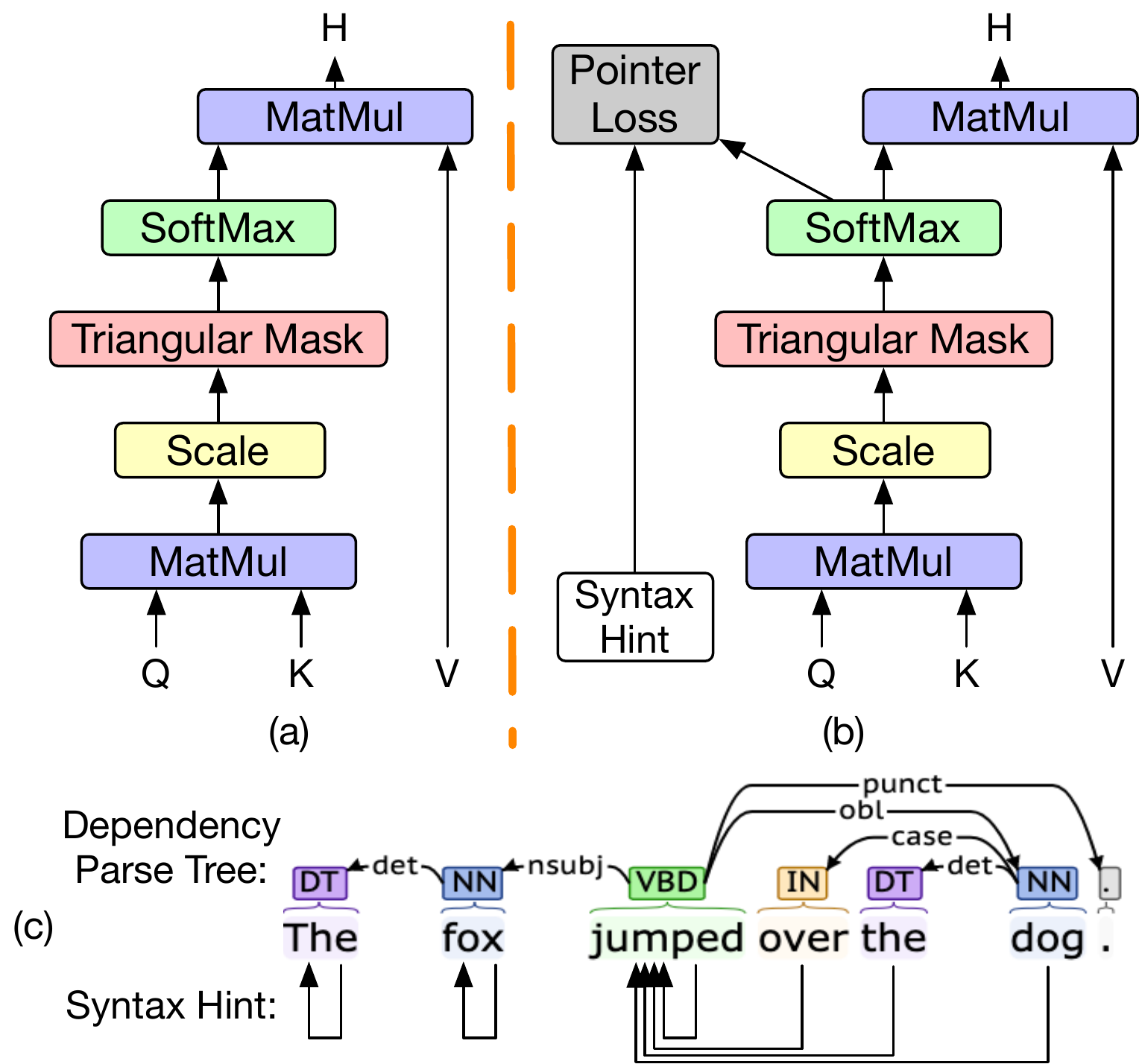}\\
    \caption{Comparison between the  conventional self-attention and syntax-guided self-attention module. (a) self-attention, (b) self-attention with syntax hint, (c) the heuristic way of obtaining syntax hint. For each token $w$, the syntax hint is another token which is the ancestor of $w$ in the dependency tree and occurred before $w$ in the sentence. If such token does not exist, then the syntax hint of $w$ is itself.}
    \label{fig:betsg}
\end{figure}
\subsection{Bird-Eye Transformers}\label{sec:bet}
Text generation models tend to use the information of historical tokens to predict the next token. However, the importance of historical tokens is not always the same. In a linguistic view, natural language is constructed by a set of \textit{syntax} rules~\cite{chomsky1956three,chomsky2014aspects},  which is a tree-like structure. So, according to the position of tokens in a syntax tree, the tokens can be roughly divided into two types: high-level tokens and low-level tokens. High-level tokens usually contain general information of the current sentence and are able to affect tokens that are far away from them, while low-level tokens can only affects nearby tokens. Therefore, paying more attention to the high-level tokens is  promising to contribute to a better prediction of  the next token in generation models, as shown in some LSTM-based works~\cite{shen2018ordered,sha2018jointly}.

We would like to  take one step further to propose a novel architecture, called bird-eye transformer (BET). This transformer architecture  refines the self-attention weights to encourage it to focus on more informative historical tokens. 

\paragraph{Syntax-guided BET}
Since syntax features of the current token usually provide useful hints to the next token, we propose to use some syntax hints to guide the attention weights changing towards the ``high-level'' tokens, which is named as BET(SG). The guidance of syntax hints is conducted by a pointer loss as is shown in Fig.~\ref{fig:betsg}(b). The syntax hints are directly taken from the dependency parse tree. If $x_{t+1}$ is the token to be predicted, then we will take its nearest ancestor node in the dependency parse tree which also occurs in the left of $x_{t+1}$ in the sentence as the syntax hint of the current token $x_t$. If no ancestor node occurs in the left of $x_{t+1}$, then the syntax hint is the current token itself. Here, this syntax hint is the ``high-level'' information w.r.t the  token to be predicted. 


The input format of the syntax hint is an one-hot vector for each token, the ``1'' lies in the position where the syntax hint located. Assume that these one-hot vectors (length $n$) for the $n$ tokens are $y_{s}\in\mathbb R^{n\times n}$, the pointer loss of each transformer block is:

\begin{equation}
L_p = \sum y_s\log A
\end{equation}

The pointer loss of each transformer block are added together and multiplies with a hyperparameter $\lambda_p$, then it is added into the final loss function.

\begin{figure}[!t]
    \centering
    \includegraphics[width=\linewidth]{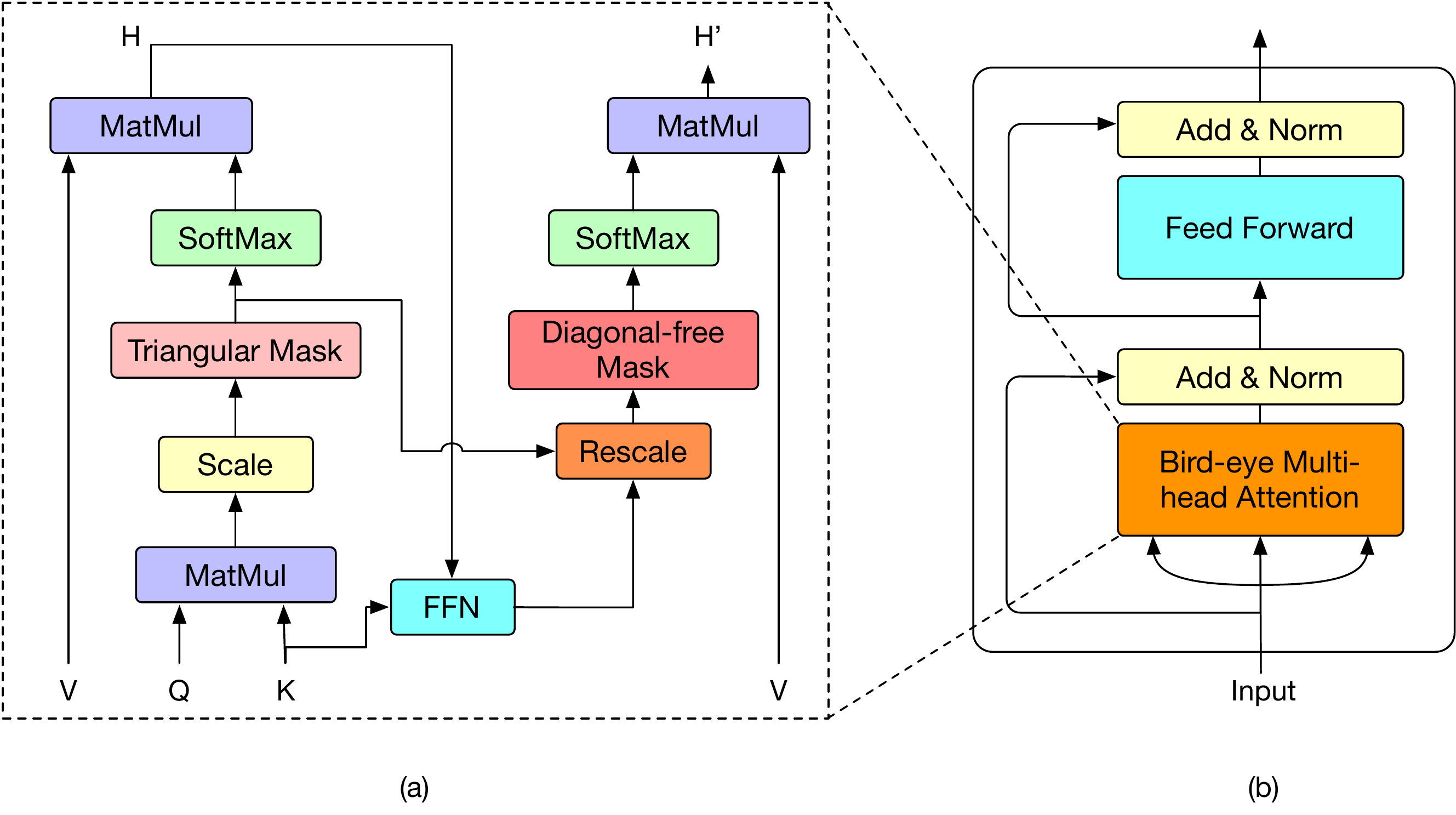}
    \caption{Simple illustration of syntax-guidance-free bird-eye transformer: (a) bird-eye attention, and  (b) bird-eye transformer.}
    \label{fig:arch}
\end{figure}

\paragraph{Syntax-guidance-free BET}
This architecture tend  to induce syntax hints by itself without external signals, which is named as BET(SF) as  shown in Fig.~\ref{fig:arch}. The main difference between BET(SF) and the standard transformer lies in two places:(1) we use the bird-eye information to reweight the self-attention, which encourages to attend to more informative historical tokens. The details are described in Figure~\ref{fig:arch}, and  (2) we add a diagonal mask to the self-attention matrix.

Intuitively, the output of self-attention collects more high-level information than the input, so it can help to decide which word of the input is a high-level word. Given the input $X$, we can get the result of self-attention $H$ as follows:
\begin{equation}
\begin{aligned}
     &Q=XW_Q,\quad K=XW_K,\quad V=XW_V, \\
   &M_{dp} = \texttt{Masked-MatMul}(Q,K),\\
   &A=\texttt{Softmax}(M_{dp}), \quad H=V^\top A
   \end{aligned}
\end{equation}
where \texttt{Masked-MatMul} uses the triangular mask to make sure that each token cannot attend to future tokens, $M_{dp}$ is the dot-product matrix, and $H$, $Q$, and $K$ are all in size $n\times d$. This process is also clearly stated in Fig.~\ref{fig:arch}~(b). 

Then, we use the hidden layer $H$ and the key $K$ to decide which word is the high-level word:
\begin{equation}
   R =  \texttt{Sigmoid}(w^\top([H,K])), 
\end{equation}
where $w\in\mathbb{R}^{2d\times 1}$ is a trainable parameter vector, ``$[\cdot,\cdot]$'' represents the concatenation of two tensors, and $R\in\mathbb{R}^{d}$ is a probability vector encoding whether the corresponding word is a high-level word. 

Finally, we use the probability vector $R$ to reweight the dot-product matrix, and recalculate the self-attention matrix as follows:
\begin{align}
M' &= M_{dp} * R,  \\
A'&=\texttt{Softmax}(M'),\\
  H' &= V^\top A'.
\end{align}
After the rescaling operation, the resulting feature matrix is expected to contain word-level information. $H'$ is the output of our BET's self-attention module. The BET's attention module also can be extended to multi-head attention and can be an alternative to the standard multi-head attention as is shown in Figure~\ref{fig:arch}(c). Since the main change of BET lies in the self-attention module, we can also stack multiple BET layers together. 
Compared to conventional transformers, the only parameter that we bring to the BET is the vector $w$ when deciding the high-level words. This parameter vector does not require too much additional memory capacity compared to the large amount of parameters in standard transformers.

\paragraph{Diagonal-Free Mask}
In a transformer layer, the residual connection~\cite{he2016deep} is able to directly feed the input to the next layer. Therefore, in the text generation model, when predicting the $(i+1)$-th token, the representation of the $i$-th token is already directly sent to the layer after self-attention. As a result, it is not necessary for the  self-attention module to attend to the current token anymore.


So, we propose to add a diagonal-free mask after the bird-eye rescaling operation, which directly mask out all the diagonal elements in the dot-product matrix before apply the \texttt{Softmax} operation to obtain the self-attention matrix. Since the first token only has one token (itself) to attend, the first row of the self-attention matrix is not to be masked out.  As shown in Fig.~\ref{fig:arch}~(b), the diagonal mask is conducted before the final \texttt{Softmax}  module. 

\section{Experiments}\label{sec:analysis}
In this section, we first introduce the datasets used in the experimental analysis sections, and we then use ablation studies to discuss the disadvantages of self-attention. Finally, we show the performance of the proposed BET model.

\subsection{Datasets and Experimental Settings}\label{sec:data}
According to our observation, the disadvantages of self-attention affect the performance of the transformer's decoder part. So, we use two text generation tasks, namely, machine translation and language modeling, to analyze the self-attention mechanism. 
The machine translation datasets are IWSLT 2014 German-English (De-En)~\cite{cettolo2016iwslt} and WMT 2017 English-German (En-De)~\cite{ondrej2017findings}, evaluated by BLEU score~\cite{papineni2002bleu}. The language model datasets are WikiText-2~\cite{merity2016pointer}, Wiki-103~\cite{merity2016pointer}, and Enwiki8~\cite{mahoney2011large}.
The evaluation metric is perplexity~\cite{brown1992estimate} for word-level datasets (WikiText-2, Wiki-103) and bits-per-character (BPC)~\cite{graves2013generating} for character-level datasets (Enwiki8).Detailed experiment settings are listed as follows. 
  
\paragraph{Machine Translation.}
We use two machine translation datasets {(under license CC-BY-SA)}: IWSLT 2014 German-English (De-En)~\cite{cettolo2016iwslt} and WMT 2017 English-German (En-De)~\cite{ondrej2017findings}. The evaluation metric is the BLEU score~\cite{papineni2002bleu}.

For German-English (De-En) machine translation, we use the same train/valid/test data splitting as previous\footnote{\url{http://www.statmt.org/wmt14/translation-task.html}}. We have $153$K  parallel sentence pairs for training, $7$k for validation, and $7$k for testing. We use BPE~\cite{sennrich2015neural} to get subword vocabularies. Then, we get  a shared source-target vocabulary of more than $10$K tokens. We develop our model based on the Fairseq\footnote{\url{https://github.com/pytorch/fairseq.git}}~\cite{ott2019fairseq} repository. We use the transformer-based encoder-decoder architecture by \newcite{vaswani2017attention} as our basic model, which is called ``iwslt14.tokenized.de-en'' in Fairseq. The hidden layer size and word embedding dimension are set to $512$. We also use 6  transformer layers for encoder and decoder. The batch size is set to 22,  Adam~\cite{kingma2014adam} is used for optimization with $\beta_1=0.9$ and $\beta_2=0.98$, and the learning rate is updated during training using the method by  \newcite{vaswani2017attention}. 

For English-German (En-De) machine translation, there are $1.9$M  parallel sentence pairs for training, $2$k for validation, and $3$k for testing. We use the same model  with previous task. After BPE segmentation, we   combine the vocabulary of English and German together as is consistent with \newcite{vaswani2017attention}. The final  vocabulary size is $25,860$. The number of encoder and decoder layers are $6$. The other settings are the same as for De-En.

\paragraph{Language Modeling.}
We use three datasets (under licence CC-BY-SA) for language modeling: WikiText-2~\cite{merity2016pointer}, Wiki-103~\cite{merity2016pointer}, and Enwiki8~\cite{mahoney2011large}. WikiText-2~\cite{merity2016pointer} is a small word-level dataset, which  contains 
$2$M training tokens and  a vocabulary size of $33$k. Wiki-103~\cite{merity2016pointer} is a large word-level dataset with a lot of long-distance dependencies. This is good for revealing the disadvantages of self-attention. There are $103$M tokens and $28$K articles for training in Wiki-103. The average length of articles is $3.6$K tokens. Enwiki8~\cite{mahoney2011large} is a character-level dataset, which  contains $100$M bytes of raw Wikipedia text. There are $205$ unique characters in the Enwiki8 dataset. 
The evaluation metric is perplexity~\cite{brown1992estimate} for word-level datasets and bits-per-character (BPC)~\cite{graves2013generating} for character-level datasets.
The dimensions of the transformers' hidden layer and word embedding are  $300$ for WikiText-2 and WikiText-103, and $512$ for Enwiki8. We use Adam~\cite{kingma2014adam} with learning rate $0.001$ for optimization.

\subsection{Transformers' Self-Attention Analysis}\label{sec:preexp}

To prove that the current token has taken too much attention weight which should belong to other historical tokens, we computed and compared the averaged self-attention weight values of the $i$-th token paid on itself and on other historical tokens.  We conduct such kind of  analysis  on the  machine translation and the language modeling task, and report the results  in Table~\ref{tab:attn}.
\begin{table*}[t]
    \caption{Quantitative analysis of self-attention. These results are collected from the converged   6-layer  transformers for the two tasks. For the machine translation task, we collect the decoder part. ``CA'' means current attention weight, ``HA($\mu$)'' and ``HA($\sigma$)'' means the average value and standard deviation of previous attention weights, and ``Ratio'' is the ratio between CA and HA($\mu$): $CA/HA$.  We use the IWSLT De-En dataset for machine translation evaluation,  and WikiText-2 for language modeling evaluation in this table. }
    \label{tab:attn}
    
    \centering
    \resizebox{0.6\linewidth}{!}{
    \begin{tabular}{c|cccc|cccc}
    \toprule[1.0pt]
          Layers& \multicolumn{4}{c|}{1} & \multicolumn{4}{c}{2}  \\
         \midrule[0.5pt]
            & CA & HA($\mu$)& HA($\sigma$) & Ratio & CA & HA($\mu$)& HA($\sigma$) & Ratio\\
         \midrule[0.5pt]
       MT &0.2964 & 0.0435 &0.0126  &6.81   & 0.0592   & 0.0475 &0.0034  &1.25  \\
        LM &0.0628 &0.0192&0.0131 &3.27    & 0.0584 & 0.0193 &0.0103  & 3.02   \\
        \midrule[1.0pt]
         Layers& \multicolumn{4}{c|}{3} &  \multicolumn{4}{c}{4} \\
         \midrule[0.5pt]
          & CA & HA($\mu$)& HA($\sigma$) & Ratio & CA & HA($\mu$)& HA($\sigma$) & Ratio\\
        \midrule[0.5pt]
        MT &0.0762 & 0.0466&0.0053& 1.64 &0.0544 & 0.0479  &0.0046 & 1.14 \\ 
        LM &0.0541  &0.0194 &0.0098&2.79 &0.0546 &0.0194 &0.0143&2.81\\
         \midrule[1.0pt]
          Layers&  \multicolumn{4}{c|}{5} &  \multicolumn{4}{c}{6}\\
         \midrule[0.5pt]
          & CA & HA($\mu$)& HA($\sigma$) & Ratio & CA & HA($\mu$)& HA($\sigma$) & Ratio\\
         \midrule[0.5pt]
        MT & 0.0629   & 0.0473  &0.0048 & 1.33 &0.0754 &0.0464 &0.0051& 1.63\\
       LM  & 0.0939 &0.0186&0.0157 &5.05 & 0.0664 &0.0192 &0.0124& 3.46  \\ 
    \bottomrule[1.0pt]
    \end{tabular}
    }
    
\end{table*}

In Table~\ref{tab:attn}, we have three metrics for analysis: CA, HA, and Ratio, which are defined as follows:
\begin{itemize}
    \item Current Attention (CA): When predicting the $(i+1)$-th token, self-attention uses the $i$-th token as the feature collector, so CA means the  attention weight of the $i$-th token to itself, which lies in the diagonal of the self-attention matrix. The number of CA in Table~\ref{tab:attn} is the average of all the diagonal weights in the  self-attention matrix throughout the test set.
    \item Historical Attention (HA): When predicting the $(i+1)$-th token, HA means the  attention weight of the $i$-th token to all the historical previous tokens (the $0\sim(i-1)$-th), which lies in the  lower triangular part of the self-attention matrix. We compute the average and standard deviation of all the lower triangular  self-attention matrix throughout the test set and record them in Table~\ref{tab:attn}.
    \item Ratio: This is the ratio between CA and HA. If this number is large, it means that the $i$-th token pays too much attention on itself.
\end{itemize}

According to Table~\ref{tab:attn}, of all the six layers in language modeling, the $i$-th token is focusing on itself (we call this phenomenon ``self-attending'' in this paper) about three times more than focusing on  historical tokens when predicting the $(i+1)$-th token.  Especially in the fifth layer, this ratio is even greater than 5. Also, in machine translation tasks, the CA/HA ratio of the first layer is extremely high, which achieves nearly $7$. This fact shows that the current self-attention mechanism allows that each token attends to themselves too much, which is not good for taking advantage of historical tokens.


\subsection{Transformers' Ablation Test Settings}
In this section, we introduce the experimental settings for testing the importance of diagonal attention values and lower triangular attention values. The former tends to discover the effect when the attention values on the current token are distributed to the historical tokens, while the latter  discovers the effect when we strengthen or weaken the distinct between the attention values  on the historical tokens.

\paragraph{The Effect of Diagonal Attention Values.}
To further investigate possible ways to improve the performance of self-attention, we make three small changes to the current self-attention mechanism, and test their effect to the performance on the two tasks of machine translation and language modeling. The three small changes and their combination are illustrated as follows:

\begin{itemize}[leftmargin=*]
\item Reduced/Magnified Diag: For weakening/strengthening the attention weights of ``self-attending'', we first multiply $20\%$/$200\%$ to the diagonal elements of the dot product matrix. Then, we conduct the \texttt{Softmax} operation.
    \item DiagFreeMask: the diagonal-free mask will mask out all the diagonal elements in the self-attention matrix but the first one. The first diagonal element will not be masked out, because the first token can only attend to itself.
    
\end{itemize}

\begin{table*}[!t]
    \caption{Overall performance comparison. ``$\uparrow$'' (resp., ``$\downarrow$'') means higher (resp., lower) is better. For machine translation, the evaluation metric is BLEU~\cite{papineni2002bleu}. For word-level language modeling (WikiText-2 and WikiText-103), it is perplexity, and for character-level language modeling (Enwiki8), it is BPC (byte-per-character). ``*''~means significantly outperforms the transformer baseline in the first line (with $p\,{<}\,0.05$ in Stu\-dent's t-test). The results of BET(SG) are underlined because BET(SG) has external syntax information as input.}
    \label{tab:overall}
    \centering
    \resizebox{\linewidth}{!}{
    \begin{tabular}{lccccccc}
    \toprule[1.0pt]
         & \multicolumn{2}{c}{Machine Translation} & \multicolumn{3}{c}{Language Modeling} \\
         \midrule[0.5pt]
         &  De-En ($\uparrow$) &   En-De ($\uparrow$) &   WikiText-2  ($\downarrow$) &   WikiText-103 ($\downarrow$) &  Enwiki8 ($\downarrow$)\\
    \midrule[0.5pt]
    Transformer &   35.26 & 27.55  & 54.53 &  24.27 &   1.164 \\
    Transformer (Reduced Diag) &  35.39 & 27.64  &  54.22 &  24.18 & 1.158*\\
    Transformer (Magnified Diag) & 35.02&	26.89&	56.73&	25.86 & 1.187\\
    Transformer $+$ DiagFreeMask &  35.42* & 27.67  &  54.08* &  24.05* & 1.155*\\
    \midrule[0.5pt]
    Hopfield & 33.70    & 25.44&54.09*  &26.45 & 1.201\\
    \newcite{yang2018modeling} &35.39&	27.64&	54.48&	24.21 &	1.160\\
    \newcite{zhao2019explicit}& 35.51	&27.84&	54.39&	24.16&	1.159\\
    Transformer-XL &-	&-&	54.39&	24.00&	1.161\\
   Routing Transformer &-	&-&53.81	 & 23.79  &	1.158    \\
    \midrule[0.5pt]
    BET(SG)      &  \uline{36.04}     &  \uline{28.43}  & \uline{51.66}& \uline{23.34} & \uline{1.147}\\
    BET(SF)      &\textbf{35.85}*   &\textbf{28.03}*  & \textbf{52.97}*& \textbf{23.76}*& \textbf{1.153}*\\
    BET(SF) $-$ DiagFreeMask   &35.67*    & 27.89& 53.50*& 23.98*& 1.156* \\
    \bottomrule[1.0pt]
    \end{tabular}
    }
\end{table*}

\subsection{Baselines for BET}
We conducted an experimental comparison with the following decoder-only baselines, all of which have made some improvements on the self-attention module:
\begin{itemize}[leftmargin=*]
    \item Hopfield Network~\cite{hopfield2007hopfield}: We use exactly the same architecture in \newcite{ramsauer2020hopfield}, which reveals the link between Hopfield networks and transformers, and integrates Hopfield networks into back-propagation-based deep learning methods. \newcite{ramsauer2020hopfield} uses the associative memory theory~\cite{radhakrishnan2020overparameterized} to refine the self-attention matrix by minimizing an energy function. In practice, this minimization process is conducted by iteratively update the self-attention matrix to achieve a stationary state. We directly integrate the code of Hopfield network layers\footnote{\url{https://github.com/ml-jku/hopfield-layers}} into our code.
    \item \newcite{yang2018modeling}: This method tries to add some Gaussian bias to the self-attention to capture useful local context. 
    \item \newcite{zhao2019explicit}: This method uses top-k sparse self-attention to concentrate on the most contributive tokens. 
    \item Transformer-XL~\cite{dai2019transformer}: This method is proposed to learn long-term dependency by a  segment-level recurrence in the hidden states. We run the released code while keep the hyperparameters the same to our settings for comparison.
    \item Routing Transformer~\cite{Roy2021EfficientCS}: This method tries to alleviate the computation cost of the self-attention module by using k-means clustering to avoid  attending to content unrelated to the query. To make the results comparable, we directly changed their hyperparameters in the source code\footnote{\url{https://github.com/google-research/google-research/tree/master/routing\_transformer}} to make them the same as our experiment settings, and rerun the experiments on our datasets.

\end{itemize}
We use the same experimental settings as in Section~\ref{sec:preexp} for a fair comparison. Our method is labeled 
with  ``BET(SG)'' and ``BET(SF)'' in the following experiments. 
For ablation tests, we remove DiagFreeMask from each BET architecture to show the change in performance.




\subsection{Overall Performance and Ablation Study}

The overall performance is shown in  Table~\ref{tab:overall}.  
We can see that, with the help of syntax hints, BET(SG) obtains the highest performance. 
When we let the model induce the syntax hints by itself, our proposed syntax-guidance-free BET method (BET(SF)) has also outperformed all the baseline methods, although not so high as BET(SG).  This demonstrates the effect of our bird-eye mechanism.

\begin{figure*}[!ht]
\centering
    \resizebox{\linewidth}{!}{
    \begin{tabular}{cc}
         \includegraphics[width=0.5\linewidth]{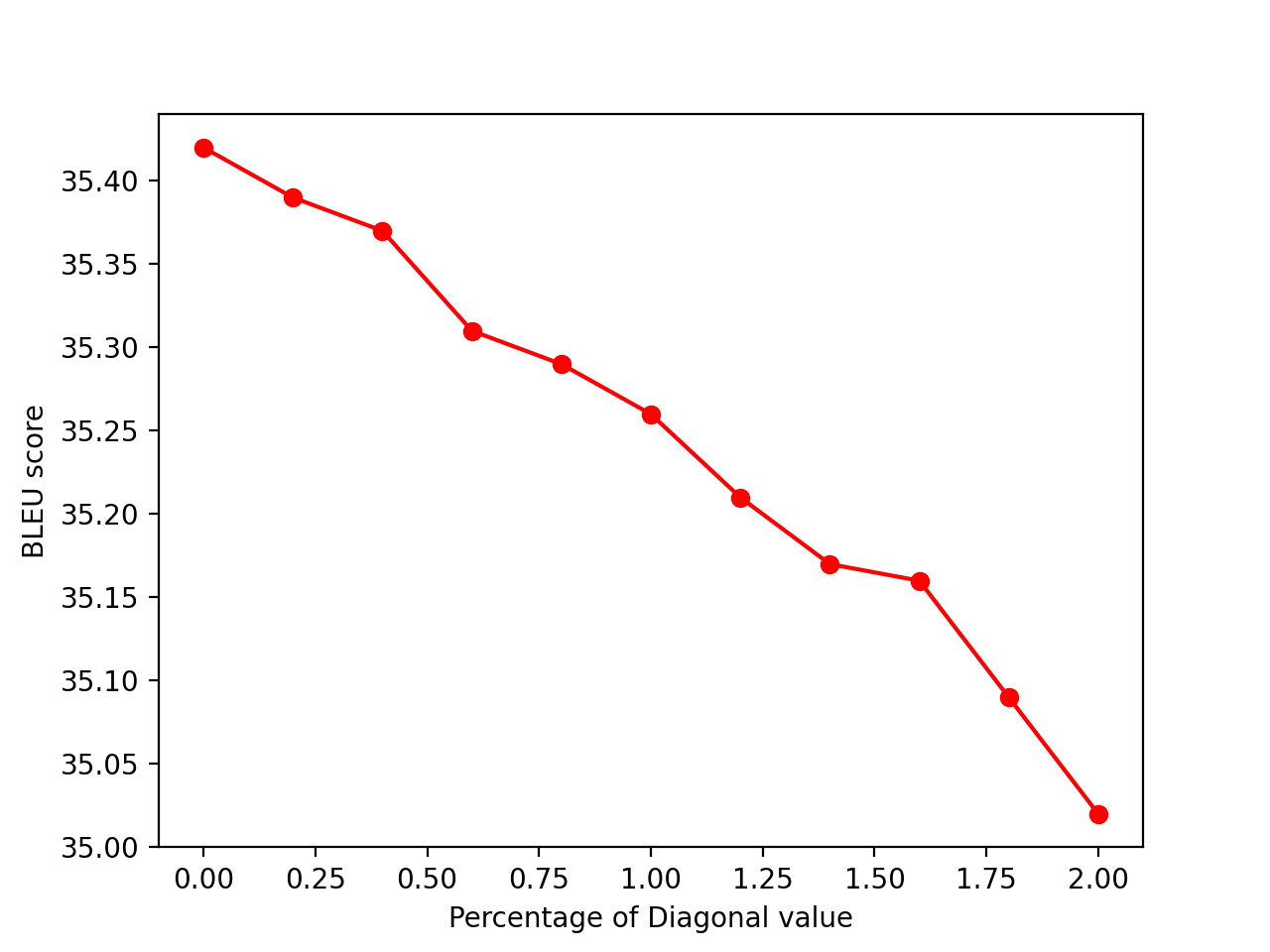}&\includegraphics[width=0.5\linewidth]{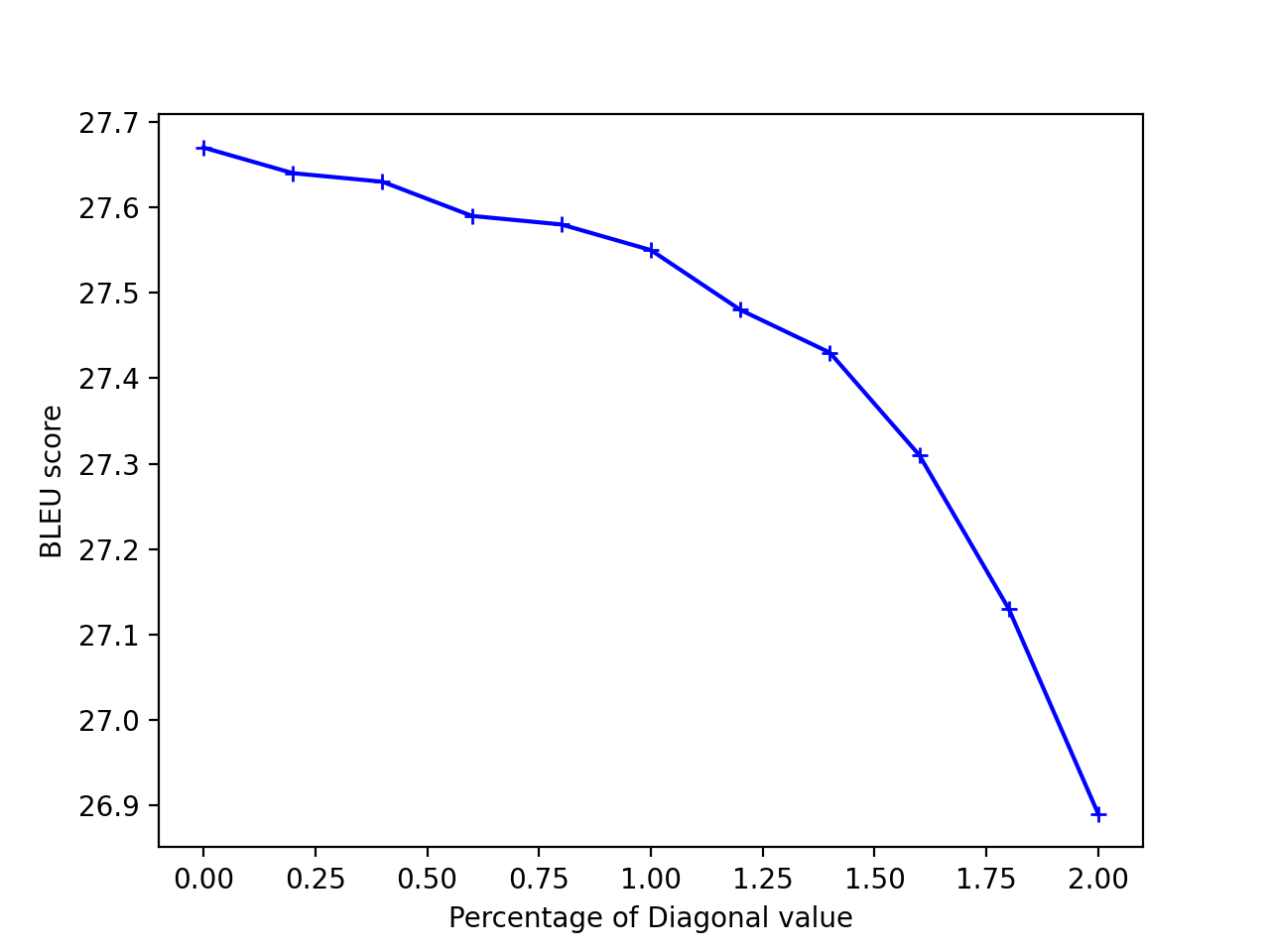}\\
         (a) & (b)\\
         \includegraphics[width=0.5\linewidth]{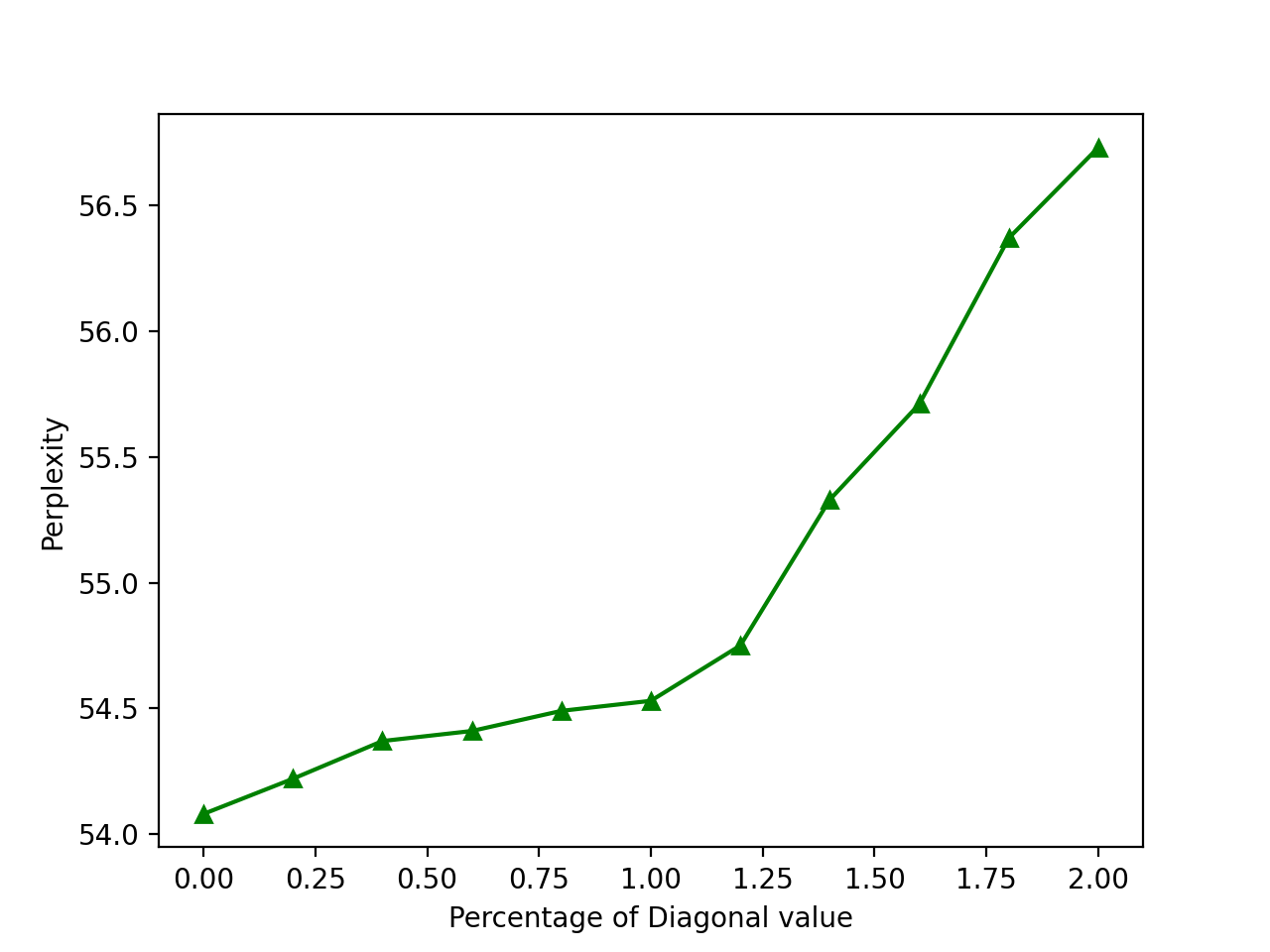}&\includegraphics[width=0.5\linewidth]{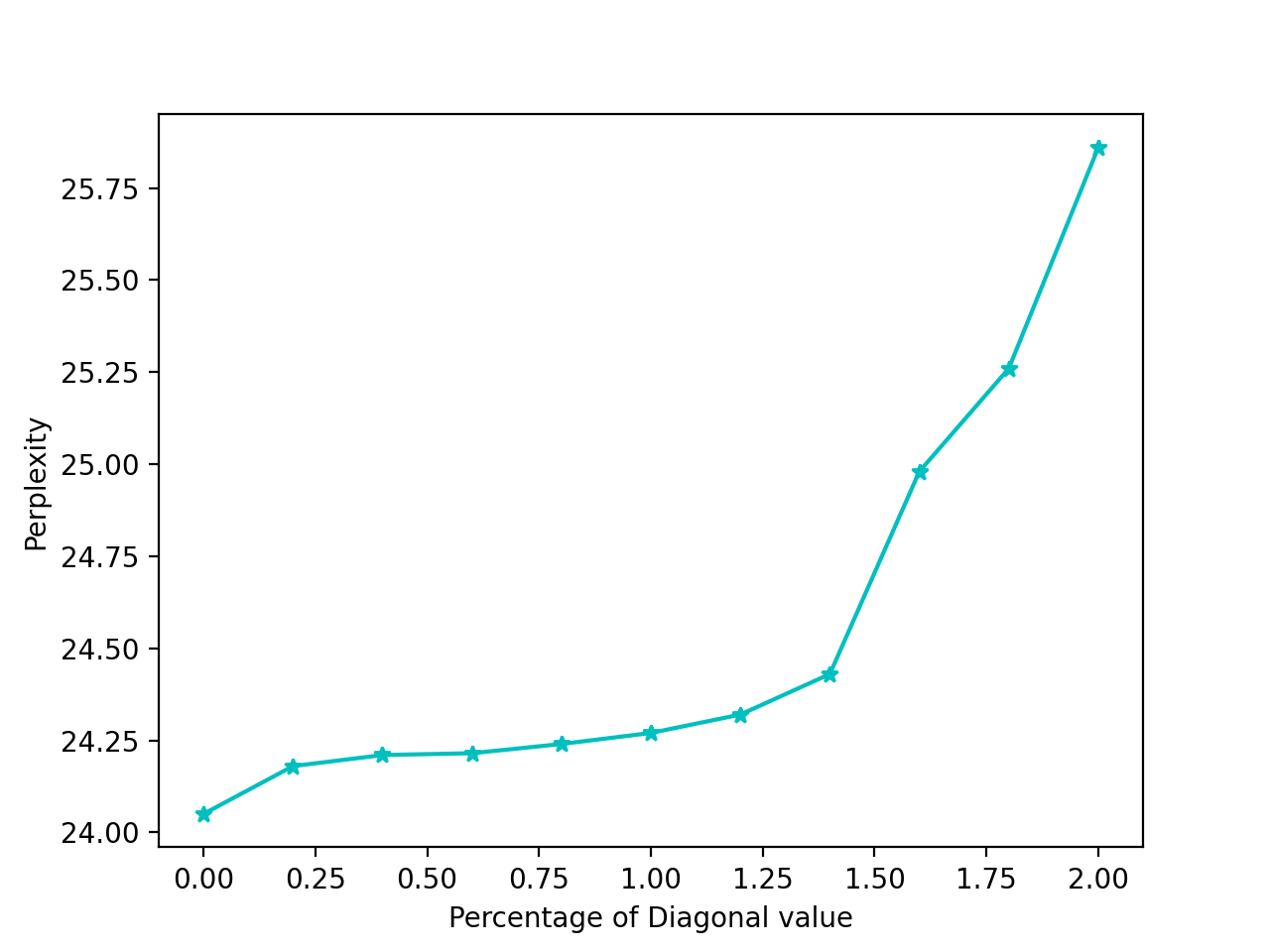}\\
         (c) & (d)\\
    \end{tabular}
    }
    \caption{When we first discount the diagonal elements of the dot-product matrix, then conduct \texttt{softmax}, the models' performance and the discount rate shows negative correlation. So, the less we keep the diagonal elements of the dot-product matrix, the better the performance will become.}
    \label{fig:disc}
\end{figure*}

Especially, we found that  DiagFreeMask is also very useful in our BET(SF) architecture.  
For the ablation tests for diagonal attention values. The results are listed as Table~\ref{tab:overall}. We can see that, after we mask out the diagonal elements of the dot-product matrix, the performance of the five tasks increased significantly, as the historical tokens obtained more attention  after the diagonal elements are masked out, and they contributed more to the prediction of the $(i+1)$-th token. Similarly, when we discount the  diagonal elements to $20\%$, the performance also improved slightly. The changing curve of the BLEU scores (for machine translation tasks) and the perplexity (for the language model tasks) according to the discount rate in shown in Figure~\ref{fig:disc}.

\begin{figure*}[!h]
    \centering
    \resizebox{\linewidth}{!}{
    \begin{tabular}{ccc}
         \includegraphics[width=0.33\linewidth]{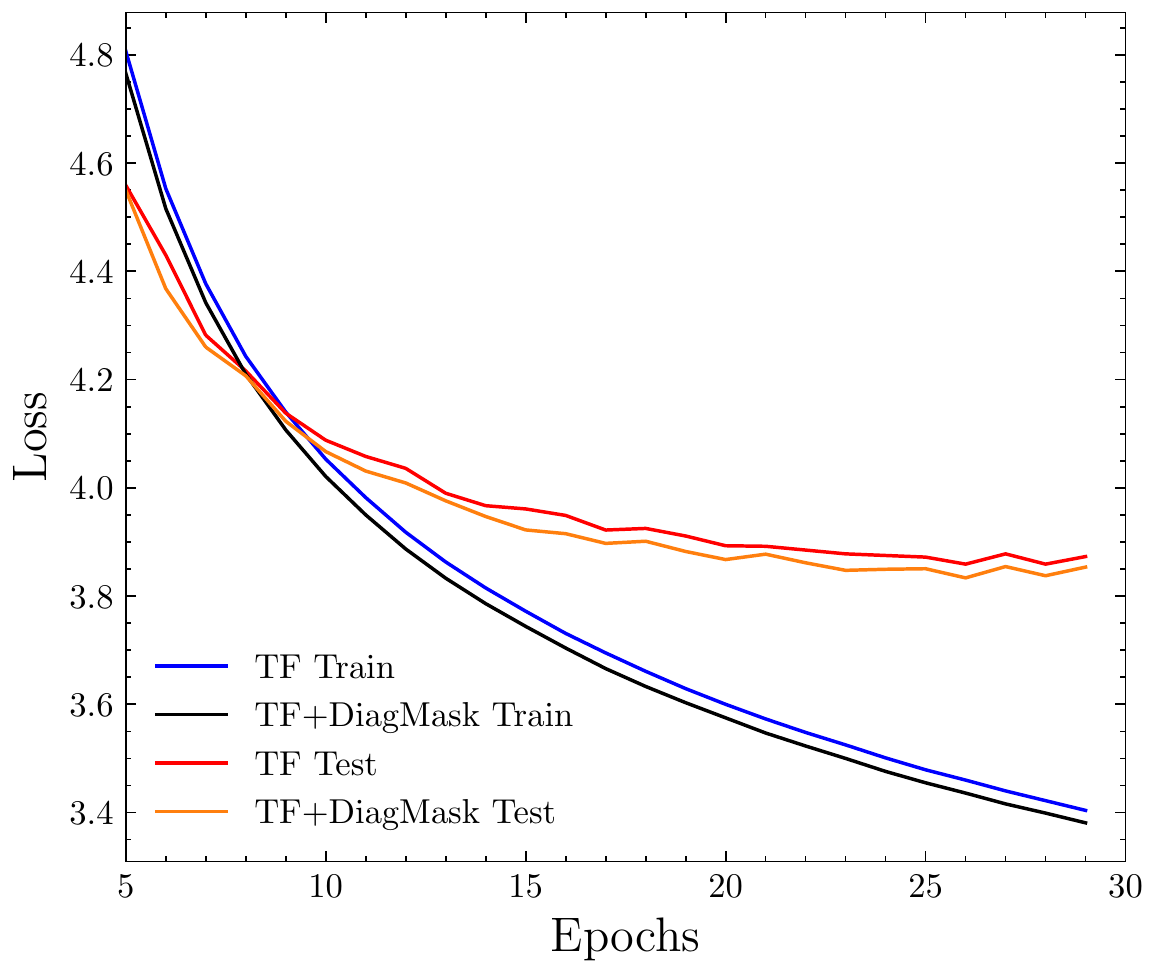}&\includegraphics[width=0.33\linewidth]{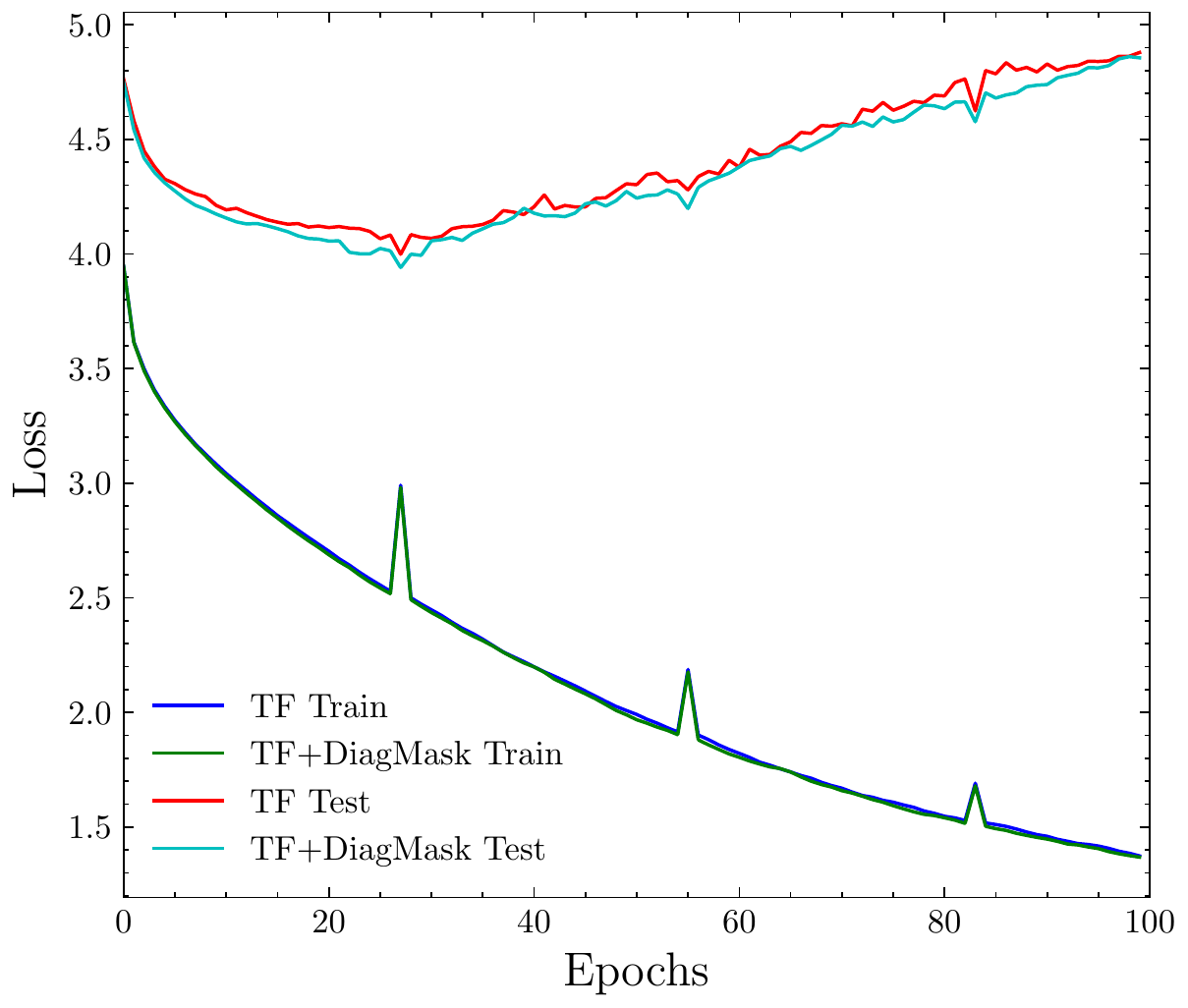}&\includegraphics[width=0.33\linewidth]{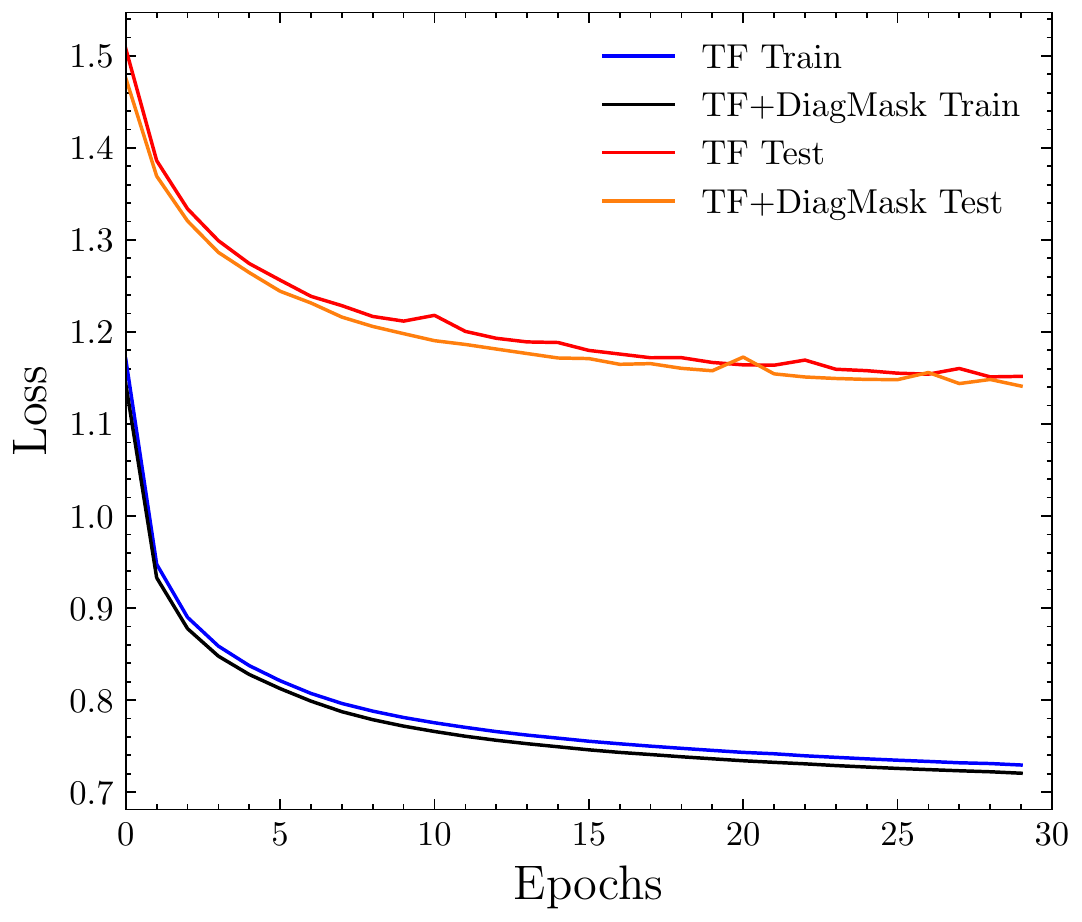}\\
        (a) & (b)& (c)
    \end{tabular}
    }
    \caption{Loss curve comparison before and after we add DiagFreeMask to the self-attention matrix: (a)~De-En machine translation, (b) WikiText-2, and (c) Enwiki8.}
    \label{fig:dmcurve}
\end{figure*}

After we removed  DiagFreeMask, the perplexities and BPCs slightly increased, which means that the language model is getting more inaccurate without DiagFreeMask. A comparison of loss curves  before and after we add DiagFreeMask to the self-attention matrix is shown in  Figure~\ref{fig:dmcurve}.

\begin{figure*}[!t]
    \centering
    \resizebox{\linewidth}{!}{
    \begin{tabular}{ccc}
         \includegraphics[width=0.33\linewidth]{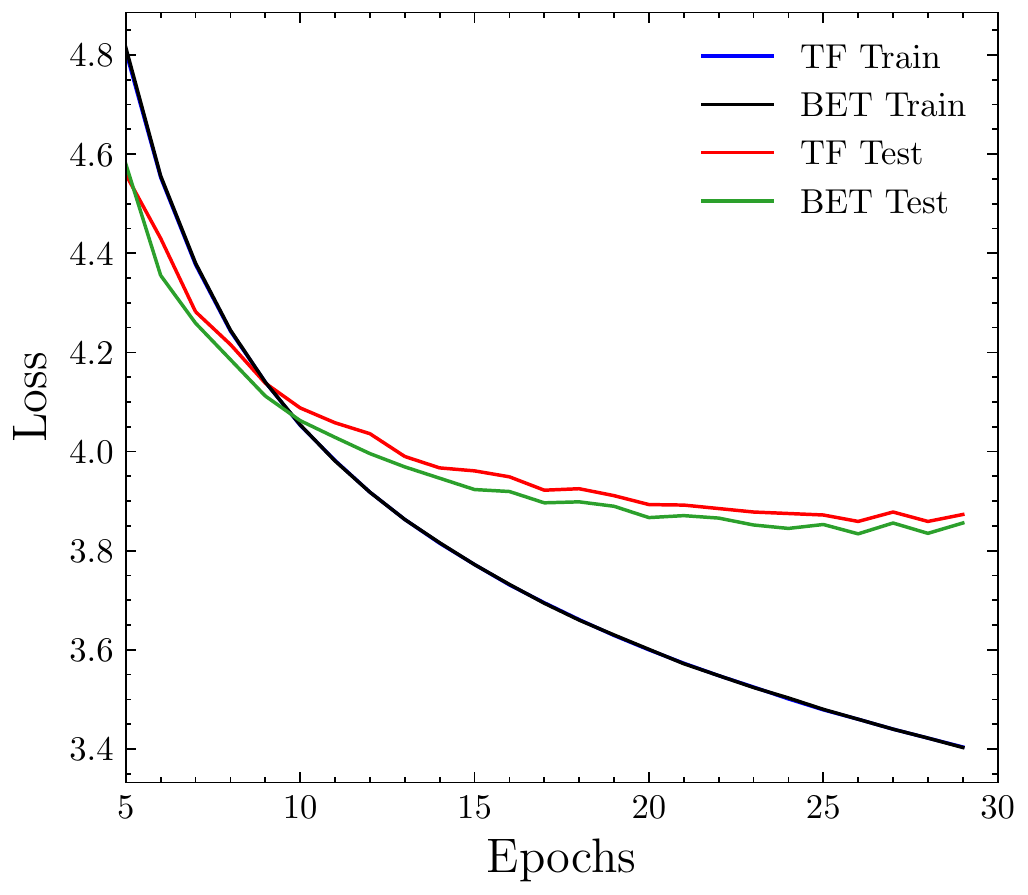}&\includegraphics[width=0.33\linewidth]{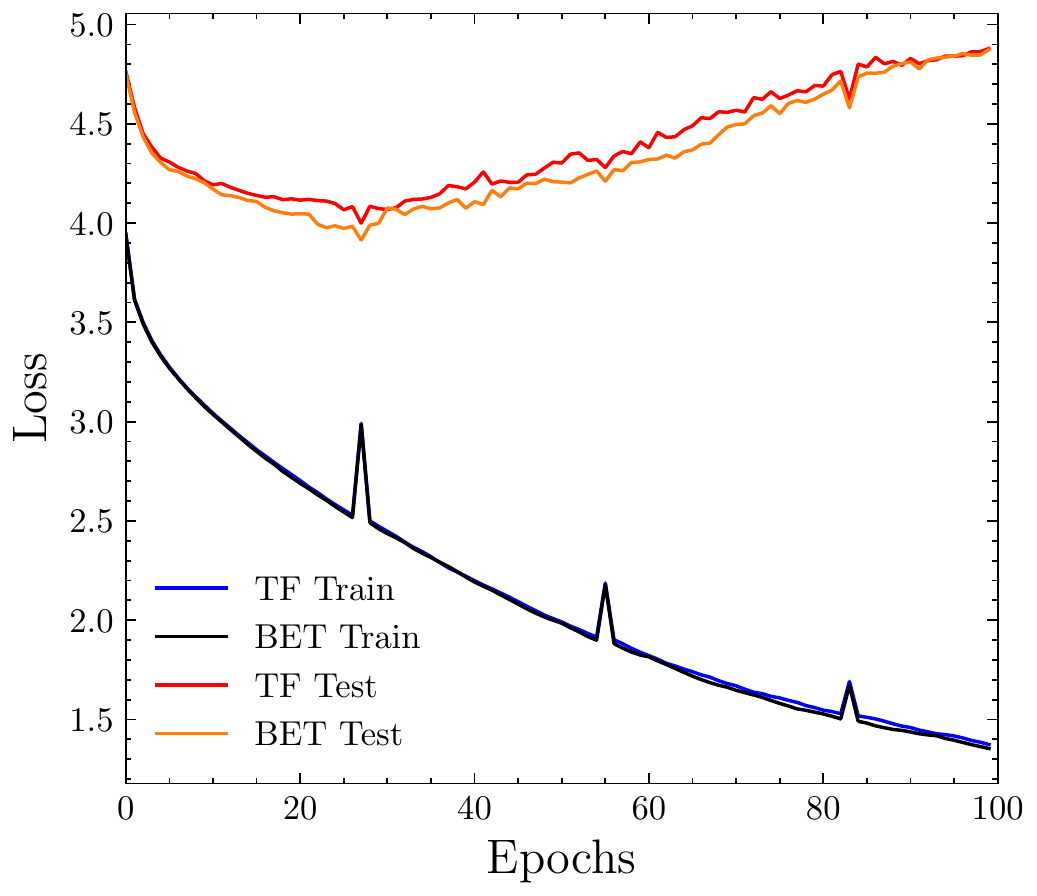}&\includegraphics[width=0.33\linewidth]{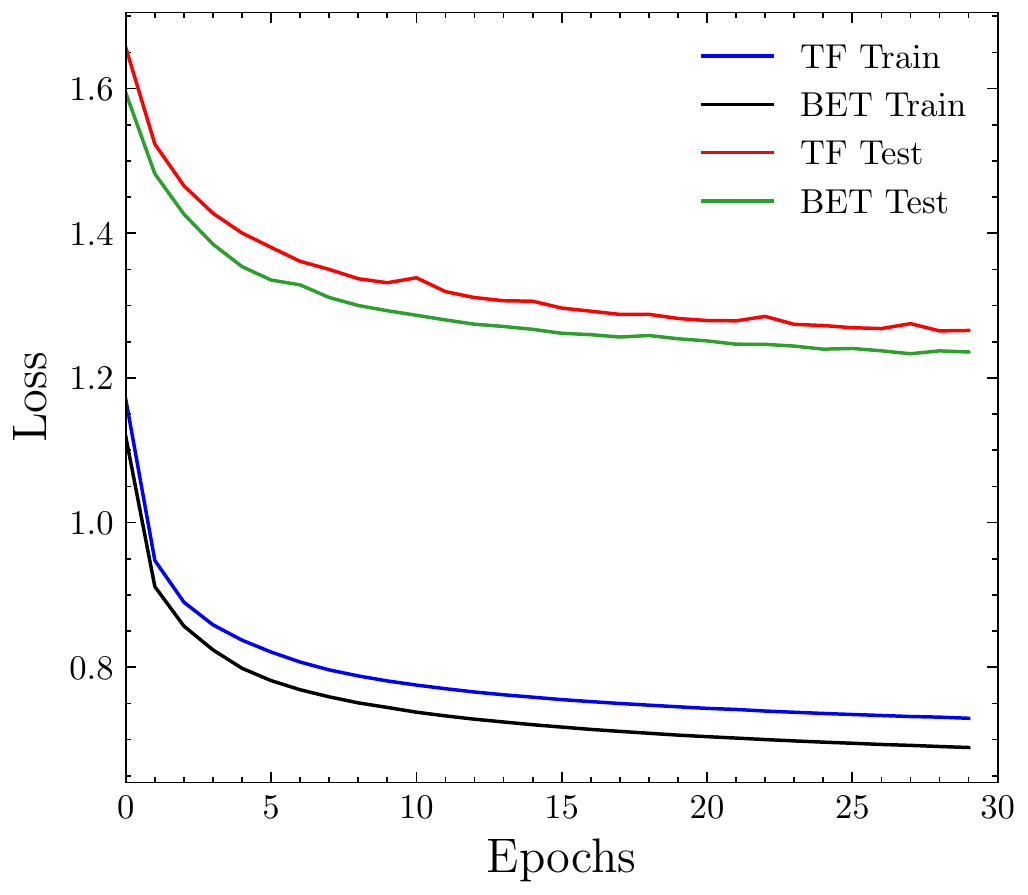}\\
        (a) & (b)& (c)
    \end{tabular}
    }
    \caption{Loss curve comparison of BET and Transformer: (a)~De-En machine translation, (b)~WikiText-2, and (c) Enwiki8.}
    \label{fig:betcurve}
\end{figure*}

Also, the comparison of the training curves for three of the datasets (Fig.~\ref{fig:betcurve}) shows a very clear  advantage of BET(SF).


 After we set the hyperparameters of all the baseline methods to  the same (including the number of layers, number of heads, dimension of hidden layers), we found that our proposed method has outperformed all the competitive methods.  The Hopfield network takes advantage of associative memories to store input patterns (texts or images), which is an initial step of integrating neuroscience into backpropagation-based deep learning. This allows Hopfield networks to outperform transformers on tasks where large associative memories are needed, e.g. multiple-instance learning tasks~\cite{ramsauer2020hopfield}. So, Hopfield network is more powerful in encoding instead of decoding. \newcite{yang2018modeling} and \newcite{zhao2019explicit} tried to use Gaussian bias or top-k sparse self-attention to focus on the tokens which have already obtained more attention weights. In contrast, our proposed method tends to focus  more on syntax-related tokens, so that the amended attention module is more informative. Transformer-XL did not specially design the self-attention module,  while routing transformers tend to focus on the tokens which can be clustered together with the current token, this is also less informative than the syntax-related tokens. Therefore, our proposed method is able to obtain a higher performance.

\subsection{Case Study}\label{sec:casestudy}
We still need to answer the question that whether the syntax-free BET can take advantage of more important tokens. 
We  listed the top attended tokens in all the six BET(SF) or transformer layers in Table~\ref{tab:casestudy}. We can see that in higher layers (the fourth to sixth layer), BET(SF) can focus on many relevant tokens when predicting the red token. For example, in the fourth layer, when predicting ``iron'', BET(SF) paid the most attention on ``warship'' while the other transformers are focusing on the previous word ``by''. Obviously, the token ``warship'' contributes more than the token ``by'' when predicting the token ``iron''. However, at the same time, in lower layers, like the second layer, the tokens focused by both BET(SF) and the other transformers may be not so relevant to the token to be predicted. This indicates that BET(SF) also can benefit from more stacked layers.
Note that in layer 3, the token ``attack'' indeed relates very closely to the previous word ``heart'', which is accurately attended by the other transformers. Although, in the attention matrix of BET(SF), the token ``heart'' is not attended, our BET(SF) architecture will not lose the information of the token ``heart'', because it can be complemented by the residual connection in BET(SF).
\begin{table*}[!t]
    \caption{Comparison of attended tokens between BET(SF) and transformers. The task is language modeling using WikiText-2. The token in \textcolor{red}{red} is the token to be predicted at the current time step. The layer number indicates the exact transformer layer where the attention values are extracted. }
    \label{tab:casestudy}
    \centering
    \resizebox{\linewidth}{!}{
    \begin{tabular}{lp{12cm}ll}
    \toprule[1.0pt]
   Layer & Sentence   & Method &  Top 3 attended tokens\\
    \midrule[0.5pt]
    1 & \multirow{2}{12cm}{the increasing violence in derry and elsewhere led to increasing \textcolor{red}{speculation} that  internment without trial would be introduced\ldots }& BET &    led, increasing, elsewhere \\\cmidrule{3-4}
 & &  Transformer   &  derry, the, increasing \\
    \midrule[0.5pt]
   2 & \multirow{2}{12cm}{manila 's healthcare is also provided by private corporations , private hospitals that \textcolor{red}{operates} in the city are the manila\ldots} & BET &   by, also, corporations \\\cmidrule{3-4}
 & &  Transformer   & by, also, private \\
    \midrule[0.5pt]
    3&  \multirow{2}{12cm}{samuel  suffered a heart attack four days after his beating , on 17 july he suffered a further heart \textcolor{red}{attack} and died \ldots} & BET &   suffered, further, he, \\\cmidrule{3-4}
 & &  Transformer   & heart, [period], he \\
  \midrule[0.5pt]
    4& \multirow{2}{12cm}{an ironclad is a steam propelled warship protected by \textcolor{red}{iron} or steel armor plates used in the early part of the\ldots} & BET &    warship, propelled, protected\\\cmidrule{3-4}
 & &  Transformer   & by, protected, propelled \\
    \midrule[0.5pt]
   5&  \multirow{2}{12cm}{on september 15 , 1999 , american beauty opened to the public in limited release at three \textcolor{red}{theaters} in los angeles and \ldots} & BET &    open, beauty, public \\\cmidrule{3-4}
 & &  Transformer   & september, 15, beauty \\
    \midrule[0.5pt]
  6 &  \multirow{2}{12cm}{congress previously held office at the old congress building , in 1972 , due to declaration of martial law , \textcolor{red}{congress}  was dissolved .} & BET &   declaration, congress, held \\\cmidrule{3-4}
 & &  Transformer   & in, [comma], old \\
    \bottomrule[1.0pt]
    \end{tabular}
  }
  \end{table*}

\section{Related Work}

After transformers were proved useful in machine translation~\cite{vaswani2017attention,tay2020efficient},  a large number of transformer variants were created. Most of them are focusing on approximating the quadratic cost self-attention matrix by a low-cost method. These models can be divided into four categories according to the computation of attention.
The first category sparsifies self-attention to decrease the computational cost. Among them, \newcite{qiu2020blockwise} and \newcite{parmar2018image} chunk the input sequence into several blocks, and only compute self-attention between blocks. Sparse Transformer~\cite{child2019generating} and  Longformer~\cite{beltagy2020longformer} consider strided attention patterns. Compressed Attention~\cite{liu2018generating} further uses strided convolution to compress the self-attention matrix. Also,  Axial Transformer~\cite{ho2019axial} combines a bunch of sparse self-attention patterns together for a better coverage of the original self-attention. Transformer-XL~\cite{dai2019transformer} further connects multiple segments and chunks by recurrence mechanism.
The second category learns to split the input sequence into chunks. For example, Reformer~\cite{kitaev2020reformer} and Routing Transformer~\cite{roy2021efficient} use a hash-based similarity measure and k-means clustering  to cluster input tokens into chunks.  The Sinkhorn Sorting Network~\cite{tay2020sparse} even learns to sort blocks of the input sequence. 
The third category uses an external memory to access all tokens at the same time, e.g., Set Transformers~\cite{lee2019set} and ETC~\cite{ainslie2020etc}. This method is similar to the parameter attention~\cite{sukhbaatar2019augmenting}. Big Bird~\cite{zaheer2020big} is build based on ETC, so it also leverages global memories.
The fourth category is low-rank methods, which assumes that the self-attention matrix can be obtained by the  multiplication of multiple low-rank matrices. Representative researches include Linformer~\cite{wang2020linformer}, Performer~\cite{choromanski2020masked}, and kernel-based methods~\cite{katharopoulos2020transformers}.

Many works also integrate tree structures into transformers. Tree transformers~\cite{wang2019tree} add a constituency prior to encourage the self-attention to learn whether two tokens belong to one span, thus a hierarchical structure can be learned by a multi-layer model. Variants of position encoding~\cite{shiv2019novel} can also help to learn tree structures. 
Although our bird-eye rescaling mechanism borrows some inspiration from the natural language's hierarchical structure, our main goal is to recognize high-level words in  historical tokens (like syntax-related tokens)  and let them  help to predict future tokens instead of learning the whole syntax tree.


Recently, methods~\cite{ramsauer2020hopfield} based on associative memories~\cite{radhakrishnan2020overparameterized,feldman2020neural,krotov2016dense,krotov2020large,marullo2021boltzmann} were shown to be related to self-attention. Associative memories~\cite{le2020self,chatterjee2018learning,wang2018associative} are a psychology concept, which is the ability to learn and memorize the relationships of unrelated items. Usually, they are applied in neuroscience-oriented approaches, and optimized by minimizing an energy function.
Hopfield networks~\cite{ramsauer2020hopfield} are a good example of integrating associative memories into  backpropagation-based neural networks. In Hopfield networks,
an energy function is minimized via an iterative updating of self-attention.

\section{Conclusion}

In this paper, we use a series of  solid experiments to show the disadvantage of the self-attention architecture in transformers. We find that the current self-attention architecture has paid too much attention weights on the diagonal element  of the self-attention matrix, while fails to provide specific attention to ``high-level'' historical token information. So, we propose a novel architecture of transformers: syntax-free bird-eye transformers (BET-SF), which is able to find syntax clues and pay more attention on syntax-related historical tokens. In the experiment analysis, we found that the BET with external syntax information (BET-SG) achieved the best performance. Although  the syntax-free BET (BET-SF) did not contain any external syntax information, it still significantly outperformed  standard transformer architectures as well as many baseline methods on all datasets of two tasks (machine translation and language modeling). 



\bibliography{cite}
\bibliographystyle{acl_natbib}

\end{document}